\documentclass[11pt]{article}

\usepackage{colacl}
\usepackage{psfig}
\usepackage{avm}
\usepackage{pictexwd}
\usepackage{rotating}
\usepackage{supertabular}

\title{A Maximum-Entropy Partial Parser for Unrestricted Text}

\author{Wojciech Skut \and Thorsten Brants\\
        Universit{\"a}t des Saarlandes\\
        Computational Linguistics\\
        D-66041 Saarbr{\"u}cken, Germany\\
        {\tt \{skut,brants\}@coli.uni-sb.de}\\[1ex]
        {\em In Proceedings of the Sixth Workshop on Very Large
        Corpora, Montr\'eal, Qu\'ebec, 1998.}}

\def\stern{\ifmmode^{\ast}\else$^{\ast}$\fi}
\def\argmax{\mathop{\rm argmax}}
\def\N{\mbox{I\kern-.2emN}}
\def\setgray#1{
\ignorespaces}
\def\unsetgray{
\ignorespaces}

\begin{document}
\maketitle


\begin{abstract}


This paper describes a partial parser that assigns syntactic structures
to sequences of part-of-speech tags. The program uses the {\em maximum
entropy} parameter estimation method, which allows a flexible
combination of different know\-ledge sources: the hierarchical structure,
parts of speech and phrasal categories. In effect, the parser goes beyond
simple bracketing and recognises even fairly complex structures. We
give accuracy figures for different applications of the parser.

\end{abstract}



\section{Introduction}

The maximum entropy framework has proved to be a powerful modelling
tool in many areas of natural language processing. Its applications
range from sentence boundary disambiguation \cite{Reynar:Ratnaparkhi:97} to
part-of-speech tagging \cite{Ratnaparkhi:96}, parsing
\cite{Ratnaparkhi:97} and machine translation
\cite{Berger:DellaPietra:DellaPietra:96}.

In the present paper, we describe a {\em partial parser} based on the
maximum entropy modelling method.  After a synopsis of the maximum
entropy framework in section \ref{sec:entropy}, we present the
motivation for our approach and the techniques it exploits (sections
\ref{sec:chunk} and \ref{sec:parsing}). Applications and results are
the subject of the sections \ref{sec:appl} and \ref{sec:results}.

\section{Maximum Entropy Modelling}
\label{sec:entropy}

The expressiveness and modelling power of the maximum entropy approach
arise from its ability to combine information coming from different
knowledge sources. Given a set $X$ of possible histories and a set $Y$
of futures, we can characterise events from the joint event space
$X,Y$ by defining a number of {\em features}, i.e., equivalence
relations over $X \times Y$. By defining these features, we express
our insights about information relevant to modelling. 

In such a formalisation, the maximum entropy technique consists in
finding a model that (a) fits the empirical expectations of the
pre-defined features, and (b) does not assume anything specific about
events that are not subject to constraints imposed by the features.
In other words, we search for the maximum entropy probability
distribution $p^*$:
\[
p^{*} = \argmax_{p\in P} H(p)
\]
where $P=\{p$:$p$ meets the empirical feature expectations\} and
$H(p)$ denotes the entropy of $p$.

For parameter estimation, we can use the Improved Iterative Scaling
(IIS) algorithm \cite{Berger:DellaPietra:DellaPietra:96}, which
assumes $p$ to have the form:
\[
p(x,y)=\frac{1}{Z(x)} \cdot e^{\sum_i \lambda_i \cdot f_i(x,y)}
\]
where $f_i:X\times Y \rightarrow \{0,1\}$ is the indicator function of
the $i$-th feature, $\lambda_i$ the weight assigned to this feature,
and $Z(x)$ a normalisation constant. IIS iteratively adjusts the weights
($\lambda_i$) of the features; the model  converges to the maximum
entropy distribution.

One of the most attractive properties of the maximum entropy approach
is its ability to cope with feature decomposition and overlapping
features. In the following sections, we will show how these advantages
can be exploited for {\em partial parsing}, i.e., the recognition of
syntactic structures of limited depth.


\section{ Context Information for Parsing}
\label{sec:chunk}

An interesting feature of many partial parsers is that they recognise
phrase boundaries mainly on the basis of cues provided by strictly
local contexts. Regardless of whether or not abstractions such as
phrases occur in the model, most of the relevant information is
contained directly in the sequence of words and part-of-speech tags to
be processed.

An archetypal representative of this approach is the method described
by \newcite{Church:88}, who used corpus frequencies to determine the
boundaries of simple non-recursive NPs. For each pair of
part-of-speech tags $t_i, t_j$, the probability of an NP
boundary (`[' or `]') occurring between $t_i$ and $t_j$ is
computed. On the basis of these context probabilities, the program
inserts the symbols `[' and `]' into sequences of part-of-speech tags.

Information about lexical contexts also significantly improves the
performance of deep parsers. For instance, \newcite{Joshi:Srinivas:94}
encode partial structures in the Tree Adjoining Grammar framework and
use tagging techniques to restrict a potentially very large amount of
alternative structures. Here, the context incorporates information
about both the terminal yield and the syntactic structure built so
far.

Local configurations of words and parts of speech are a particularly
important knowledge source for lexicalised grammars. In the Link
Grammar framework \cite{Lafferty:ea:92,DellaPietra:ea:94}, strictly
local contexts are naturally combined with long-distance information
coming from {\em long-range trigrams}.

Since modelling syntactic context is a very knowledge-intensive
problem, the maximum entropy framework seems to be a particularly
appropriate approach. \newcite{Ratnaparkhi:97} introduces several {\em
contextual predicates} which provide rich information about the
syntactic context of nodes in a tree (basically, the structure and
category of nodes dominated by or dominating the current
phrase). These predicates are used to guide the actions of a parser.

The use of a rich set of contextual features is also the basic idea of
the approach taken by \newcite{Hermjakob:Mooney:97}, who employ
predicates capturing syntactic and semantic context in their parsing
and machine translation system.

\section{A Partial Parser for German}
\label{sec:parsing}

The basic idea underlying our approach to partial parsing can be
characterised as follows:

\begin{itemize}

\item An appropriate encoding format makes it possible to express all
relevant lexical, categorial and structural information in a finite
alphabet of {\em structural tags} assigned to words (section
\ref{sec:format}).

\item Given a sequence of words tagged with part-of-speech labels, a
Markov model is used to determine the most probable sequence of
structural tags (section \ref{sec:parser}).

\item Parameter estimation is based on the maximum entropy technique,
which takes full advantage of the multi-dimensional character of the
{\em structural tags} (section \ref{sec:parest}).

\end{itemize}

The details of the method employed are explained in the remainder of
this section.

\subsection{Relevant Contextual Information}
\label{sec:format}

Three pieces of information associated with a word $w_i$ are considered
relevant to the parser:
\begin{itemize}
\item the part-of-speech tag $t_i$ assigned to $w_i$
\item the structural relation $r_i$ between $w_i$ and its predecessor
$w_{i-1}$
\item the syntactic category $c_i$ of $parent(w_i)$
\end{itemize}
On the basis of these three dimensions, {\em structural tags} are
defined as triples of the form $S_i = \langle t_i,r_i,c_i\rangle$. For
better readability, we will sometimes use attribute-value matrices to
denote such tags.

\begin{figure}[h]
\begin{center}
\begin{avm}
$S_i=$ \[
TAG $t_i$\\
REL $r_i$\\
CAT $c_i$\\
\]
\end{avm}
\end{center}
\end{figure}
Since we consider structures of limited depth, only seven values of the
REL attribute are distinguished.

\bigskip

{\tt
\[
r_i = \left \{ \begin{array}{rcl}
        0 & \mbox{\rm if} & parent(w_i)=parent(w_{i-1})\\
        + & \mbox{\rm if} &parent(w_i)=parent^2(w_{i-1})\\
        ++ & \mbox{\rm if} &parent(w_i)=parent^3(w_{i-1})\\
        - & \mbox{\rm if} &parent^2(w_i)=parent(w_{i-1})\\
        -- & \mbox{\rm if} &parent^3(w_i)=parent(w_{i-1})\\
        = & \mbox{\rm if} &parent^2(w_i)=parent^2(w_{i-1})\\
        1 & \mbox{  } & \mbox{\rm else}\\
        \end{array} \right.
\]
}
\bigskip

\begin{figure}[h]
\hrule
\bigskip
\begin{center}
\psfig{file=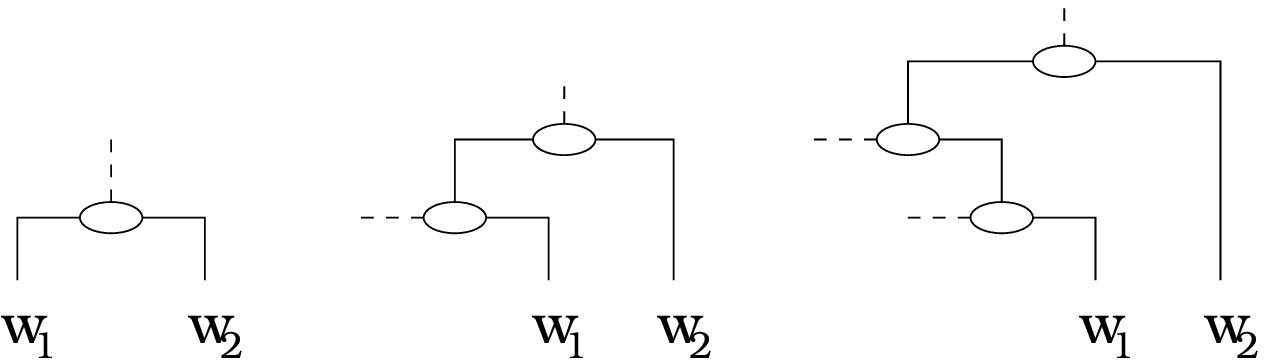,height=2.0cm,rwidth=8cm}
\end{center}
\vspace*{-3ex}
\def\h#1{\hspace*{#1}}
{\sf\h{.6em}$r_2=0$\h{3.5em}$r_2=+$\h{3.8em}$r_2=++$}\\[1ex]
\begin{center}
\psfig{file=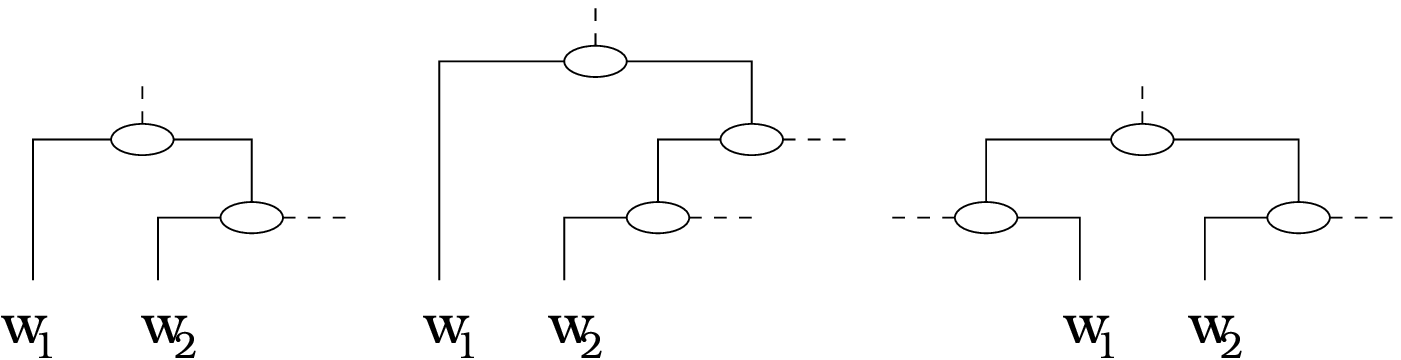,height=2.0cm,rwidth=8cm}
\end{center}
\vspace*{-4.5ex}
\def\h#1{\hspace*{#1}}
{\sf\h{.6em}$r_2=-$\h{3.5em}$r_2=--$\h{3.8em}$r_2=$ `='}\\[1ex]
\hrule
\caption{Tags $r_2$ assigned to word $w_2$}
\label{fig:str}
\end{figure}

If more than one of the conditions above are met, the first of the
corresponding tags in the list is assigned. Figure~\ref{fig:str}
exemplifies the encoding format. 

\begin{figure}[h]
\hrule
\bigskip
\begin{center}
\psfig{file=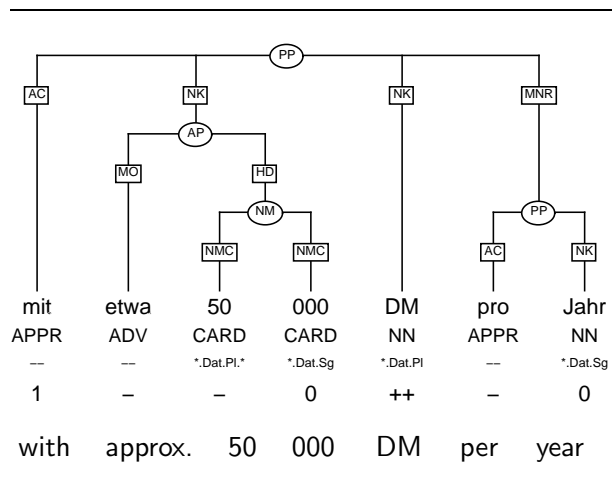,angle=-90,width=8.5cm,rwidth=8cm}
\end{center}
\vspace*{-6.5ex}
\def\h#1{\hspace*{#1}}
{\sf\h{.3em}with\h{1.2em}approx.\h{1.2em}50\h{1.2em}000\h{1.4em}DM\h{1.3em}per\h{1.3em}year}\\[1ex]
\hrule
\caption{A sample structure. The labels are explained in Appendix \ref{sec:tagsets}.}

\label{fig:samchu}
\end{figure}

These seven values of the $r_i$ attribute are mostly sufficient to
represent the structure of even fairly complex NPs, PPs and {APs},
involving {PP} and genitive {NP} attachment as well as complex
prenominal modifiers. The only {NP} components that are not treated
here are relative clauses and infinitival complements. A German
prepositional phrase and its encoding are shown in figure
\ref{fig:samchu}.

\subsection{A Markovian Parser}
\label{sec:parser} 

The task of the parser is to determine the best sequence of triples
$\langle t_i, r_i, c_i \rangle$ for a given sequence of part-of-speech
tags $\langle t_0, t_1,... t_n\rangle$. Since the attributes TAG, REL
and CAT can take only a finite number of values, the number of such
triples will also be finite, and they can be used to construct a
$2$-nd order Markov model. The triples $S_i=\langle
t_i, r_i, c_i \rangle$ are states of the model, which
emits POS tags ($t_j$) as signals.

In this respect, our approach does not much differ from standard
part-of-speech tagging techniques. We simply assign the most probable
sequence of structural tags $S= \langle S_0, S_1, ..., S_n\rangle$ to a
sequence of part-of-speech tags $T = \langle t_0, t_1, ...,
t_n\rangle$. Assuming the Markov property, we obtain:

\vbox{%
\begin{eqnarray}
        \argmax_S P(S|T) & & \hspace*{6em}
\end{eqnarray}
\vspace{-2ex}
\begin{eqnarray*}
        && = \argmax_R P(S)\cdot P(T|S) \\
        && = \argmax_R \prod_{i=1}^k P(S_i|S_{i-2},S_{i-1}) P(t_i|S_i)
\end{eqnarray*}
}
The part-of-speech tags are encoded in the structural tag (the  $t_i$
dimension), so $S$ uniquely determines $T$. Therefore, we have
$P(t_i|S_i) = 1$ if $S_i = \langle t_i,r_i,c_i\rangle$ and 0 otherwise,
which simplifies calculations.

\subsection{Parameter Estimation}
\label{sec:parest}

The more interesting aspect of our parser is the estimation of
contextual probabilities, i.e., calculating the probability of a
structural tag $S_i$ (the ``future'') conditional on its immediate
predecessors $S_{i-1}$ and $S_{i-2}$ (the ``history'').

\begin{table}[h]
\begin{center}
\begin{tabular}{cc|c}\hline
\multicolumn{2}{c|}{history} & future \\ \hline
\begin{avm}
\[CAT: $c_{i-2}$\\
  REL: $r_{i-2}$\\
  TAG: $t_{i-2}$ \]
\end{avm}
&
\begin{avm}
\[CAT: $c_{i-1}$\\
  REL: $r_{i-1}$\\
  TAG: $t_{i-1}$ \]
\end{avm}
&
\begin{avm}
\[CAT: $c_{i}$\\
  REL: $r_{i}$\\
  TAG: $t_{i}$ \]
\end{avm}
\end{tabular}
\end{center}
\end{table}


In the following two subsections, we contrast the traditional HMM
estimation method and the maximum entropy approach.
 
\subsubsection{Linear Interpolation}
\label{sec:linint}

One possible way of parameter estimation is to use standard HMM
techniques while treating the triples $S_i = \langle
t_i,c_i,r_i\rangle$ as atoms.  Trigram probabilities are estimated
from an annotated corpus by using relative frequencies $r$:
\[
r(S_i|S_{i-2},S_{i-1}) = \frac{f(S_{i-2},S_{i-1},S_i)}{f(S_{i-2},S_{i-1})}
\]
A standard method of handling sparse data is to use a linear
combination of unigrams, bigrams, and trigrams $\hat p$:
\begin{eqnarray*}
\hat p(S_i|S_{i-2},S_{i-1}) & = & \lambda_1 r(S_i) \\
                            &   & + \lambda_2 r(S_i|S_{i-1}) \\
                            &   & + \lambda_3 r(S_i|S_{i-2},S_{i-1})
\end{eqnarray*}
The $\lambda_i$ denote weights for different context sizes and sum up
to 1. They are commonly estimated by deleted interpolation \cite{Brown:ea:92}.

\subsubsection{Features}

A disadvantage of the traditional method is that it considers only full
$n$-grams $S_{i-n+1},..., S_i$ and ignores a lot of contextual
information, such as regular behaviour of the single attributes TAG,
REL and CAT. The maximum entropy approach offers an attractive
alternative in this respect since we are now free to define features
accessing different constellations of the attributes. For instance, we
can abstract over one or more dimensions, like in the context
description in figure~\ref{fig:samfeat}.
\begin{table}[h]
\begin{center}
\begin{tabular}{cc|c}\hline
\multicolumn{2}{c|}{history} & future \\ \hline
\begin{avm}
\[TAG: {\em ART}\] 
\end{avm}
&
\begin{avm}
\[TAG: {\em ADJA}\\
  REL: $0$ \] 
\end{avm}
&
\begin{avm}
\[CAT: {\em NP}\\
  REL: $0$\\
  TAG: {\em NN} \]
\end{avm}
\end{tabular}
\end{center}
\caption{A  partial trigram feature}
\label{fig:samfeat}

\end{table}

Such ``partial $n$-grams'' permit a better exploitation of information
coming from contexts observed in the training data. We say that a
feature $f_k$ defined by the triple $\langle
M_{i-2},M_{i-1},M_i\rangle$ of attribute-value matrices is {\em
active} on a trigram context $\langle S'_{i-2},S'_{i-1},S'_i\rangle$
(i.e., $f_k(S'_{i-2},S'_{i-1},S'_i)=1$) iff $M_j$ unifies with the
attribute-value matrix $M'_j$ encoding the information contained in
$S'_i$ for $j=i-2, i-1, i$.  A novel context would on average
activate more features than in the standard HMM approach, which treats
the $\langle t_i, r_i, c_i \rangle$ triples as atoms.

The actual features are extracted from the training corpus in the
following way: we first define a number of {\em feature patterns} that
say which attributes of a trigram context are relevant. All feature
pattern instantiations that occur in the training corpus are stored;
this procedure yields several thousands of features for each pattern.

After computing the weights $\lambda _i$ of the features occurring in
the training sample, we can calculate the contextual probability of a
multi-dimensional structural tag $S_i$ following the two tags $S_{i-2}$
and $S_{i-1}$:

\[
p(S_i|S_{i-2},S_{i-1})=\\ \frac{1}{Z(x)} 
\cdot e^{\sum_{i} \lambda_i \cdot f_i(S_{i-2},S_{i-1},S_i)}
\]



We achieved the best results with 22 empirically determined feature
patterns comprising full and partial $n$-grams, $n\leq 3$.  These
patterns are listed in Appendix \ref{sec:patterns}.

\begin{figure*}[t]
\hrule
\bigskip
\psfig{file=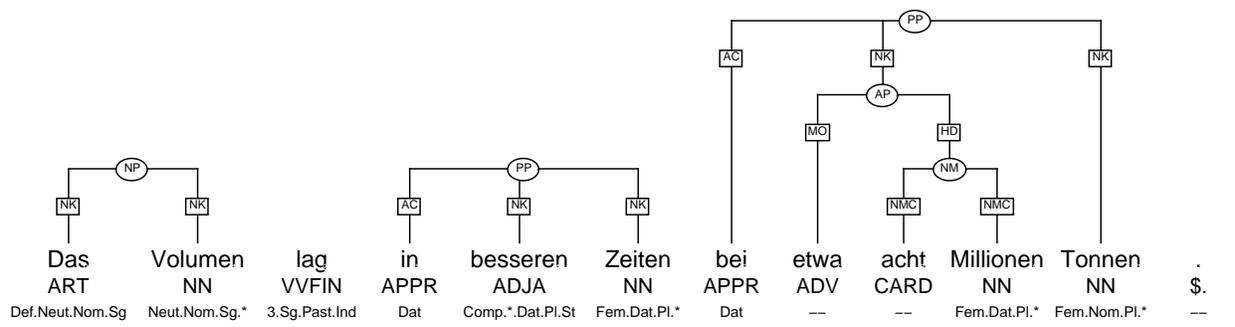,angle=-90,width=16.5cm}
\vspace*{-3ex}
\def\h#1{\hspace*{#1}}
\vspace{1.3ex}
\hrule
\caption{A chunked sentence ({\em in better times, the volume was
around eight million tons\/}). Grammatical function labels: {\sf NK}
nominal kernel component, {\sf AC} adposition, {\sf NMC} number
component, {\sf MO} modifier.}
\label{fig:ex}
\end{figure*}
\section{Applications}
\label{sec:appl}

Below, we discuss two applications of our maximum entropy parser:
treebank annotation and chunk parsing of unrestricted text. For precise
results, see section \ref{sec:results}.

\subsection{Treebank Annotation}

The partial parser described here is used for corpus annotation in a
treebank project, cf. \cite{Skut:ea:97a}. The annotation process is
more interactive than in the Penn Treebank approach
\cite{Marcus:ea:94}, where a sentence is first preprocessed by a
partial parser and then edited by a human annotator. In our method,
manual and automatic annotation steps are closely interleaved. Figure
\ref{fig:ex} exemplifies the human-computer interaction during
annotation.

The annotations encode four kinds of linguistic information: 1) parts
of speech and inflection, 2) structure, 3) phrasal categories (node
labels), 4) grammatical functions (edge labels).

Part-of-speech tags are assigned in a preprocessing step. The
automatic instantiation of labels is integrated into the assignment of
structures. The annotator marks the words and phrases to be grouped
into a new substructure, and the node and edge labels are inserted by
the program, cf. \cite{Brants:ea:97}.

Initially, such annotation increments were just local trees of depth
one. In this mode, the annotation of the {PP} {\em bei etwa acht
Millionen Tonnen\/} ({\em [at] around eight million tons\/})
involves three annotation steps (first the number phrase {\em acht
  Millionen}, then the {AP}, and the {PP}). %
Each time, the annotator highlights the immediate constituents of the
phrase being constructed.

The use of the partial parser described in this paper makes it
possible to construct the whole PP in only one step: The annotator
marks the words dominated by the PP node, and the internal structure
of the new phrase is assigned automatically. This significantly
reduces the amount of manual annotation work. The method yields
reliable results in the case of phrases that exhibit a fairly rigid
internal structure.  More than $88\%$ of all {NPs}, {PPs} and {APs}
are assigned the correct structure, including {PP} attachment and
complex prenominal modifiers.

\begin{figure*}
\hrule
\bigskip
\psfig{file=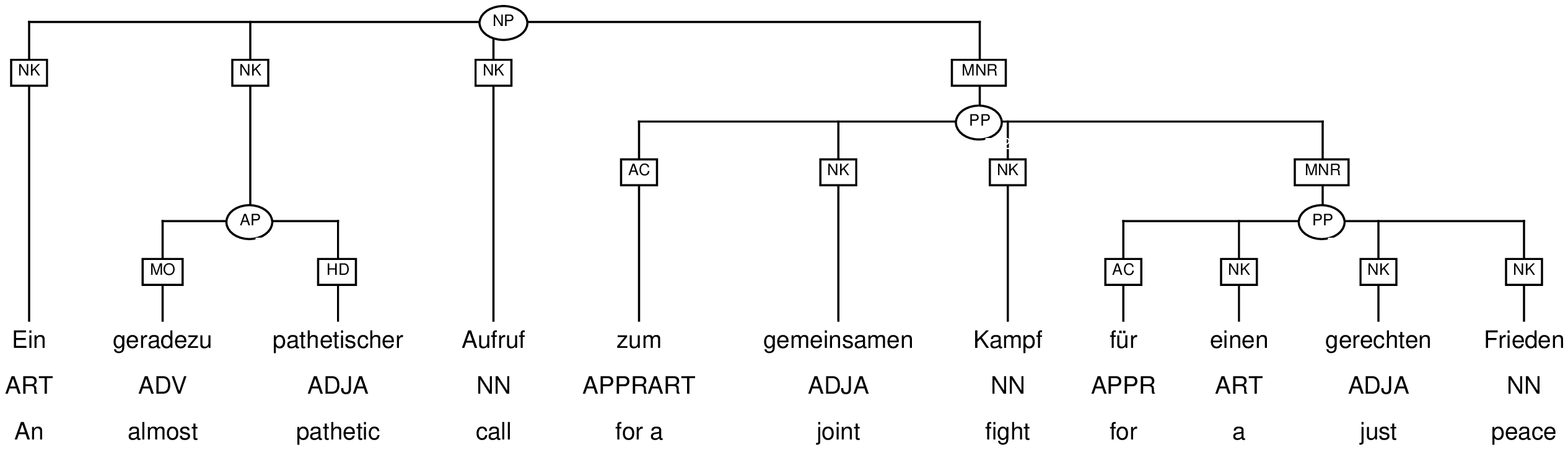,width=\textwidth,angle=-90}
\bigskip
\psfig{file=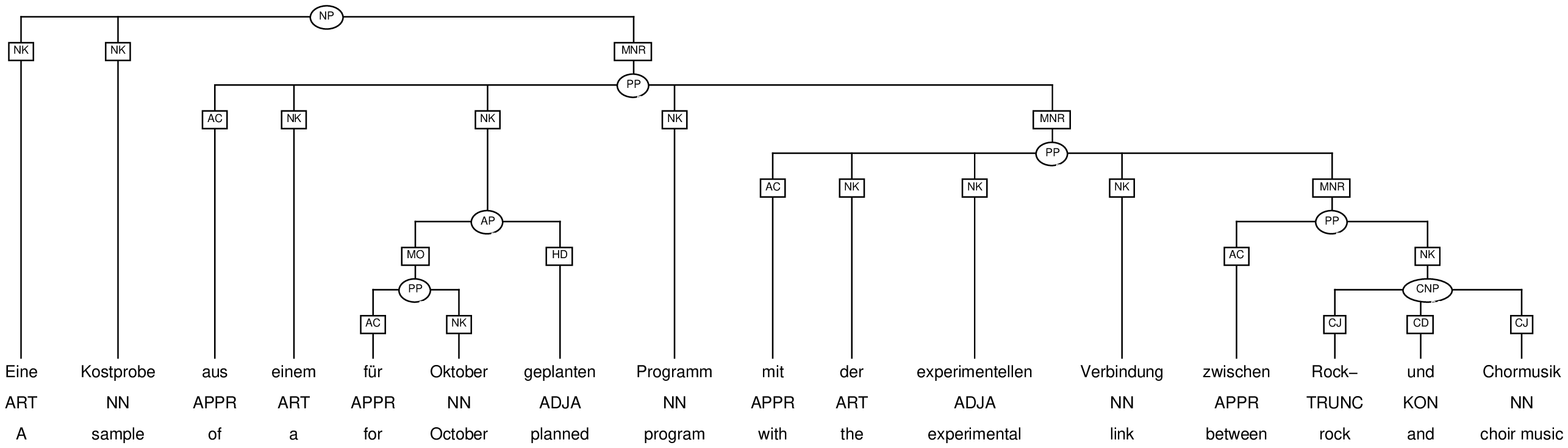,width=\textwidth,angle=-90}
\bigskip
\psfig{file=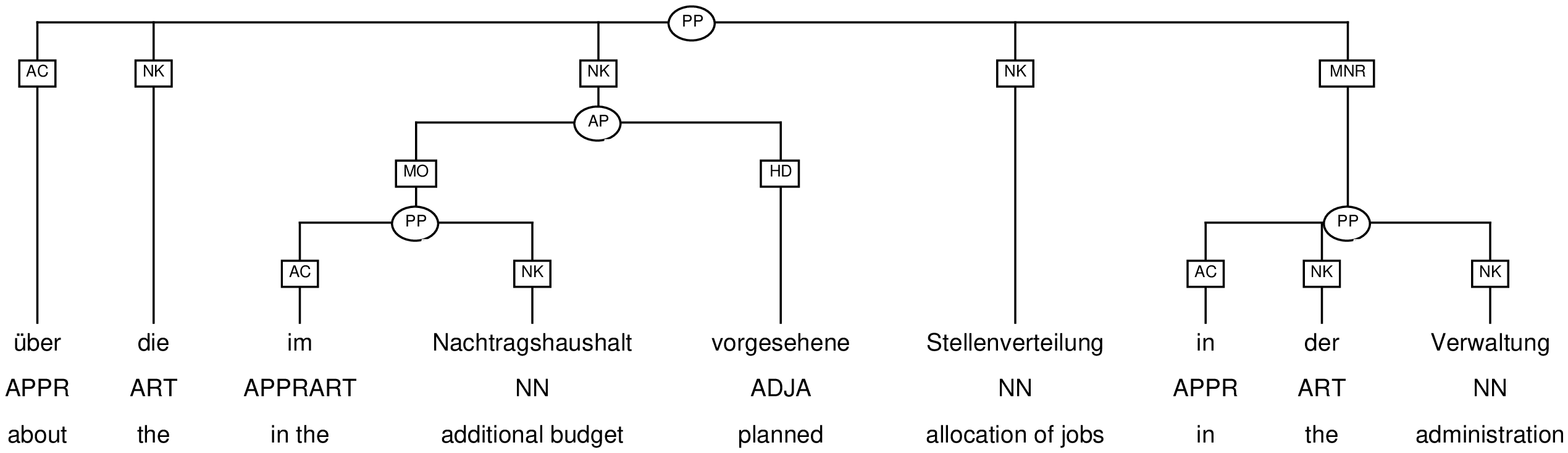,width=\textwidth,angle=-90}
\hrule

\caption{Examples of complex NPs and PPs correctly recognised by the
parser. In the treebank application, such phrases are part of larger
structures. The external boundaries (the first and the last word of
the examples) are highlighted by an annotator, the parser recognises
the internal boundaries and assigns labels.}
\label{fig:resultex}
\end{figure*}

Further examples of structures recognised by the parser are shown in
figure \ref{fig:resultex}. A more detailed description of the
annotation mode can be found in \cite{Brants:Skut:98}.

\subsection{NP Chunker}

Apart from treebank annotation, our partial parser can be used to chunk
part-of-speech tagged text into major phrases. Unlike in the previous
application, the tool now has to determine not only the internal
structure, but also the external boundaries of phrases. This makes the
task more difficult; especially for determining {PP} attachment.

However, if we restrict the coverage of the parser to the prenominal
part of the {NP/PP}, it performs quite well, correctly assigning
almost $95\%$ of all structural tags, which corresponds to
a bracketing precision of ca. $87\%$. 


\section{Results}
\label{sec:results}

In this section, we report the results of a cross-validation of the
parser carried out on the NeGra Treebank \cite{Skut:ea:97a}. The
corpus was converted into structural tags and partitioned into a
training and a testing part ($90\%$ and $10\%$, respectively). We
repeated this procedure ten times with different partitionings; the
results of these test runs were averaged.

The weights of the features used by the maximum entropy parser were
determined with the help of the Maximum Entropy Modelling Toolkit,
cf. \cite{Ristad:96}. The number of features reached 120,000 for the
full training corpus (12,000 sentences). Interestingly, tagging
accuracy decreased after after $4$--$5$ iterations of Improved
Iterative Scaling, so only $3$ iterations were carried out in each of
the test runs.

The accuracy measures employed are explained as follows.

\begin{description}
\item[tags:] the percentage of structural tags with the correct value
$r_i$ of the REL attribute,
\item[bracketing:] the percentage of correctly recognised nodes,
\item[labelled bracketing:] like bracketing, but including the
syntactic category of the nodes,
\item[structural match:] the percentage of correctly recognised
tree structures (top-level chunks only,  labelling is ignored).
\end{description}

\subsection{Treebank Application}

In the treebank application, information about the external boundaries
of a phrase is supplied by an annotator. To imitate this situation, we
extracted from the NeGra corpus all sequences of part-of-speech tags
spanned by { NPs} { PPs}, { APs} and complex adverbials. Other tags
were left out since they do not appear in chunks recognised by the
parser. Thus, the sentence shown in figure \ref{fig:ex} contributed
three substrings to the chunk corpus: {\sf ART NN}, {\sf APPR ADJA NN}
and {\sf APPR ADV CARD NN NN}, which would also be typical annotator
input. A designated separator character was used to mark chunk
boundaries.

Table \ref{table:prec} shows the performance of the parser on the
chunk corpus.


\begin{table}
\caption{Recall and precision results for the interactive annotation mode. }
\label{table:prec}
\medskip
\hrule
\begin{center}\small
\begin{tabular}{l|r|r|r|r}
measure & total & correct & recall & prec. \\ \hline
tags & 129822 &  123435  & \multicolumn{2}{c}{95.1\%} \\
bracketing & 56715 & 49715 & 87.7\% &  89.1\% \\
lab. brack. & 56715 & 47415  &   83.6\% & 84.8\% \\
struct. match & 37942 &  33450  &   88.2\%  &   88.0\% 
\end{tabular}
\end{center}
\hrule
\end{table}

\subsection{Chunking Application}

Table~\ref{table:prec2} shows precision and recall for the chunking
application, i.e., the recognition of kernel NPs and PPs in
part-of-speech tagged text. Post-nominal PP attachment is
ignored. Unlike in the treebank application, there is no pre-editing
by a human expert. The absolute numbers differ from those in 
table~\ref{table:prec} because certain structures are ignored. The total
number of structural tags is higher since we now parse whole sentences
rather then separate chunks.

In addition to the four accuracy measures defined above, we also give
the percentage of chunks with correctly recognised external boundaries
(irrespective of whether or not there are errors concerning their
internal structure). 

\begin{table}[h]
\caption{Recall and precision for the chunking application. The parser
recognises only the prenominal part of the NP/PP (without
focus adverbs such as  {\em also, only}, etc.). }
\label{table:prec2}
\medskip
\hrule
\begin{center}\small
\begin{tabular}{l|r|r|r|r}
measure & total & correct & recall & prec. \\ \hline
tags &  166995   &   158541 & \multicolumn{2}{c}{94.9\%} \\
bracketing & 51912     & 45241   & 87.2\%  &   86.9\% \\
lab. brack. & 51912  & 43813 & 84.4\%  & 84.2\% \\
struct. match & 46599  & 41422  & 88.9\%  &   87.6\% \\
ext. bounds  & 46599 & 43833  & 94.1\%   &   93.4\% 
\end{tabular}
\end{center}
\hrule
\end{table}

\subsection{Comparison to a Standard Tagger}

\begin{figure*}
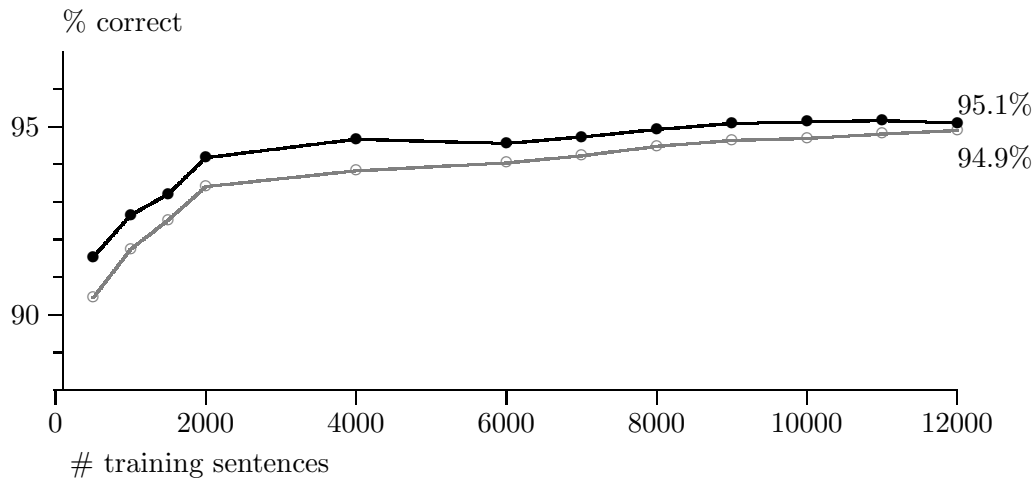

\hrule
\begin{center}
\bigskip
\hspace*{0pt}
\beginpicture
\setcoordinatesystem units <.01mm,5mm>
\setplotarea x from 100 to 12000, y from 88 to 97
\setplotsymbol({\rule{.4pt}{.4pt}})
\axis bottom ticks numbered from 0 to 12000 by 2000 /
\axis left ticks numbered 
        at 90 95 / /
\axis left ticks short
        from 88 to 96 by 1 /
\put {\# training sentences} <0mm,-10mm> [l] at 190 88
\put {\% correct} [lb] at 95 97.5
\setlinear
\setgray{.5}
\setplotsymbol({\rule{.8pt}{.8pt}})
\plot
500     90.45
1000    91.74
1500    92.52
2000    93.41
4000    93.83
6000    94.04
7000    94.23
8000    94.48
9000    94.64
10000   94.69
11000   94.81
12000   94.9
/
\multiput {$\circ$} at
500     90.45
1000    91.74
1500    92.52
2000    93.41
4000    93.83
6000    94.04
7000    94.23
8000    94.48
9000    94.64
10000   94.69
11000   94.81
12000   94.9
/
\unsetgray
\put {94.9\%} [l] at 12000 94.2
\setplotsymbol({\rule{.8pt}{.8pt}})
\plot
500     91.53
1000    92.65
1500    93.21
2000    94.18
4000    94.67
6000    94.56
7000    94.73
8000    94.93
9000    95.09
10000   95.13
11000   95.17
12000   95.1
/
\multiput {$\bullet$} at
500     91.53
1000    92.65
1500    93.21
2000    94.18
4000    94.67
6000    94.56
7000    94.73
8000    94.93
9000    95.09
10000   95.13
11000   95.17
12000   95.1
/
\put {95.1\%} [l] at 12000 95.6
\endpicture
\medskip
\end{center}
\hrule
\caption{Tagging precision  achieved by the maximum entropy parser
  (\hbox to 0pt{\hspace*{1.5mm}$\bullet$\hss}\rule[1mm]{5mm}{.8pt}) and a tagger using linear
  interpolation (\setgray{0.5}\hbox to 0pt{\hspace*{1.5mm}$\circ$\hss}\rule[1mm]{5mm}{.8pt}\unsetgray). 
  Precision is shown for different numbers of training sentences.}
\label{fig:results}
\end{figure*}

In the following, we compare the performance of the maximum-entropy
parser with the precision of a standard HMM-based approach trained on
the same data, but using only the frequencies of complete trigrams,
bigrams and unigrams, whose probabilities are smoothed by linear
interpolation, as described in section~\ref{sec:linint}. 

Figure \ref{fig:results} shows the percentage of correctly assigned
values $r_i$ of the REL attribute depending on the size of the
training corpus. Generally, the maximum entropy approach outperforms
the linear extrapolation technique by about 0.5\% -- 1.5\%, which
corresponds to a 1\% -- 3\% difference in structural match. The difference
decreases as the size of the training sample grows. For the full
corpus consisting of 12,000 sentences, the linear interpolation tagger
is still inferior to the maximum entropy one, but the difference in
precision becomes insignificant (0.2\%). Thus, the maximum entropy
technique seems to particularly advantageous in the case of sparse
data.


\section{Conclusion}

We have demonstrated a partial parser capable of recognising simple and
complex {NPs}, {PPs} and {APs} in unrestricted German text.
The maximum entropy parameter estimation method allows us to optimally
use the context information contained in the training sample. On the
other hand, the parser can still be viewed as a Markov model, which
guarantees high efficiency (processing in linear time). The program can
be trained even with a relatively small amount of treebank data; then
it can be used for parsing unrestricted pre-tagged text.

As far as coverage is concerned, our parser can handle recursive
structures, which is an advantage compared to simpler techniques such
as that described by~\newcite{Church:88}. On the other hand, the
Markov assumption underlying our approach means that only strictly
local dependencies are recognised. For full parsing, one would
probably need non-local contextual information, such as the {\em
long-range trigrams} in  Link Grammar~\cite{DellaPietra:ea:94}.

Our future research will focus on exploiting morphological and lexical
knowledge for partial parsing. Lexical context is particularly
relevant for the recognition of genitive {NP} and {PP} attachment,
as well as complex proper names. We hope that our approach will
benefit from related work on this subject,
cf. \cite{Ratnaparkhi:Roukos:94}. Further precision gain can also be
achieved by enriching the structural context, e.g. with information
about the category of the grandparent node.


\section{Acknowledgements}

This work is part of the DFG Collaborative Research Programme 378 {\em
Resource-Adaptive Cognitive Processes}, Project C3 {\em Concurrent
Grammar Processing}.

Many thanks go to Eric S. Ristad. We used his freely available Maximum
Entropy Modelling Toolkit to estimate  context probabilities.


\section*{Appendix A: Feature Patterns}

\makeatletter
\write\@auxout{\string
  \newlabel{sec:patterns}{{A}{\thepage}}}
\makeatother

Below, we give the 22 $n$-gram feature patterns used in our
experiments.

\begin{center}
\begin{tabular}{lcc|c}
\cline{2-4}
 & \multicolumn{2}{c|}{history} & future \\ 
\cline{2-4}
\begin{rotate}{90}\hspace*{-35mm}Trigram features\end{rotate}
&  $r,t,c$        &     $r,t,c$  & $r,t,c$ \\
&  $r^{sibl},t,c$ &     $r,t,c$   & $r,t,c$ \\
&  $r,c$        &     $r,c$  & $r,t,c$ \\
&  $t$          &     $r^{sibl},c$  & $r,t,c$ \\
&  $r^{sibl},t$ &     $r,t$  & $r,t$ \\
&  $r,t$        &     $r,t,c$  & $r,t$ \\
&  $r,c$        &     $r,t,c$  & $r,c$ \\
&  $r$        &     $r,t,c$  & $r$ \\
&  $r,t,c$        &     $r,t,c$  & $r,t$ \\
&  $r,t,c$        &     $r,t,c$  & $r,c$ \\
&  $t$        &     $r,t,c$  & $r,c$ \\
\hline
\begin{rotate}{90}\hspace*{-30mm}Bigram features\end{rotate}
&&  $r,t,c$    &     $r,t,c$  \\
&&  $r,t,c$    &     $r,t$  \\
&&  $r,t,c$    &     $r,c$  \\
&&  $r^{sibl},t$    &     $r,t$  \\
&&  $r,t$    &     $r,t$  \\
&&  $c$        &     $r,c$  \\
&&  $t$        &     $r,t$  \\
&&  $r$    &     $r$  \\
\hline
\begin{rotate}{90}\hspace*{-12mm}\shortstack[l]{Unigram\\features}\end{rotate}
&&&$r,t,c$  \\
&&&$r,t$ \\
&&&$r^{sibl},t$ \\
\end{tabular}
\end{center}

The symbols $r$ (REL), $t$ (TAG), and $c$ (CAT) indicate which
attributes are taken into account when generating a feature according
to a particular pattern. $r^{sibl}$ is a binary-valued attribute
saying whether the word under consideration and its immediate
predecessor are siblings (i.e., whether or not $r=0$).


\section*{Appendix B: Tagsets}

\makeatletter
\write\@auxout{\string
  \newlabel{sec:tagsets}{{B}{\thepage}}}
\makeatother

This section contains descriptions of tags used in this
paper. These are {\em not} complete lists.


\subsection*{B.1 Part-of-Speech Tags}

We use the Stuttgart-T{\"u}bingen-Tagset. The complete set is described
in \cite{Thielen:Schiller:94}.

\medskip

\begin{small}
\noindent
\begin{tabular}{ll}
{\sf ADJA}      & attributive adjective\\
{\sf ADV}       & adverb \\
{\sf APPR}      & preposition\\
{\sf APPRART}   & preposition with determiner\\
{\sf ART}       & article\\
{\sf CARD}      & cardinal number\\
{\sf KON}       & Conjunction\\
{\sf NE}        & proper noun \\
{\sf NN}        & common noun \\
{\sf PROAV}     & pronominal adverb \\
{\sf TRUNC}     & first part of truncated noun\\
{\sf VAFIN}     & finite auxiliary \\
{\sf VAINF}     & infinite auxiliary \\
{\sf VMFIN}     & finite modal verb\\
{\sf VVFIN}     & finite verb\\
{\sf VVPP}      & past participle of main verb\\
\end{tabular}
\end{small}


\subsection*{B.2 Phrasal Categories}

\begin{small}
\noindent
\begin{tabular}{ll}
{\sf AP}        & adjective phrase\\
{\sf MPN}       & multi-word proper noun \\
{\sf NM}        & multi token numeral\\
{\sf NP}        & noun phrase\\
{\sf PP}        & prepositional phrase \\
{\sf S}         & sentence \\
{\sf VP}        & verb phrase \\
\end{tabular}
\end{small}


\subsection*{B.3 Grammatical Functions}

\begin{small}
\noindent
\begin{tabular}{ll}
{\sf AC}        & adpositional case marker\\
{\sf HD}        & head\\
{\sf MO}        & modifier \\
{\sf MNR}       & post-nominal modifier\\
{\sf NG}        & negation\\
{\sf NK}        & noun kernel\\
{\sf NMC}       & numerical component\\
{\sf OA}        & accusative object \\
{\sf OC}        & clausal object \\
{\sf PNC}       & proper noun component \\
{\sf SB}        & subject \\
\end{tabular}
\end{small}

\end{document}